\documentclass[preprint,12pt]{elsarticle}

\usepackage{amssymb}
\usepackage{amsmath}
\usepackage{xcolor}
\usepackage[T1]{fontenc}
\journal{Physica A: Statistical Mechanics and its Applications}

\begin{document}

\begin{frontmatter}



\title{Enhancing Multistep Prediction of Multivariate Market Indices Using Weighted Optical Reservoir Computing}


\author[label1]{Fang Wang}
\affiliation[label1]{organization={Department of Physics, Stevens Institute of Technology},%Department and Organization
            city={Hoboken},
            postcode={07030}, 
            state={New Jersey},
            country={USA}}

\author[label2]{Ting Bu}
\affiliation[label2]{organization={Quantum Computing Inc.},%Department and Organization
            city={Hoboken},
            postcode={07030}, 
            state={New Jersey},
            country={USA}}

\author[label1,label2]{Yuping Huang}
\ead{yhuang5@stevens.edu}

%% Abstract
\begin{abstract}
%We propose and experimentally demonstrate an innovative stock index prediction method using a weighted optical reservoir computing (RC) system. We construct fundamental market data combined with macroeconomic data and technical indicators to capture the broader behavior of the stock market. We compare our optical reservoir computing prediction with traditional machine learning methods, including linear regression, decision trees, and neural network architectures such as long short-term memory. The optical RC system is enhanced by incorporating additional weights into the input features, thereby improving performance on certain benchmarks. Our results demonstrate the system's capability to effectively capture high volatility and nonlinear market behavior using limited data, showcasing its potential for real-time, parallel multi-dimensional data processing and predictions.

We propose and experimentally demonstrate an innovative stock index prediction method using a weighted optical reservoir computing system. We construct fundamental market data combined with macroeconomic data and technical indicators to capture the broader behavior of the stock market. Our approach shows significant higher performance than state-of-the-art methods such as linear regression, decision trees, and neural network architectures including long short-term memory. It captures well the market's high volatility and nonlinear behaviors despite limited data, demonstrating great potential for real-time, parallel, multi-dimensional data processing and predictions.

\end{abstract}


% \begin{highlights}
% \end{highlights}

%% Keywords

\end{frontmatter}



\section{Introduction}
\label{introduction}

Over the past decades, machine learning (ML) and deep learning have emerged as a transformative tool in the field of market prediction, particularly for predicting market indices \cite{lo2015index}. 
%An index monitors the performance of a broad asset class, such as all listed stocks to track returns on a buy-and-hold basis. 
The primary challenge to generate accurate prediction is aggregating diverse information for the benchmarks and constructing a reliable model. Techniques including regression and classification, as well as more advanced methods like deep learning and ensemble models, offer sophisticated methods for analyzing data to forecast stock and market trends, \cite{shen2012stock, leung2014machine, soni2022machine, henrique2019literature, jiang2021applications, vijh2020stock, singh2017stock, obthong2020survey, nikou2019stock, kumbure2022machine}. Especially, long short-term memory (LSTM) networks, a specialized form of recurrent neural networks (RNNs), have gained significant attention in the realm of stock prediction due to their ability to model temporal dependencies and sequence data effectively \cite{nelson2017stock, moghar2020stock, wang2018lstm, mehtab2021stock, ghosh2022forecasting, sunny2020deep, selvin2017stock}. 
%Introduced by Hochreiter and Schmidhuber in 1997\cite{hochreiter1997long}, LSTMs address the vanishing gradient problem inherent in traditional RNNs by incorporating a unique architecture of memory cells and gates (input, output, and forget gates) that regulate information flow. This structure allows LSTMs to retain relevant information over extended periods, making them particularly suitable for financial time series forecasting, where long-term dependencies and sequential patterns play a crucial role. The ability to handle non-linearities and capture intricate temporal correlations makes LSTMs a powerful tool for predicting future stock prices and trends
Together, ML models and LSTM can incorporate diverse data sources, including historical prices, technical indicators \cite{dash2016hybrid}, and macroeconomic indicators to generate comprehensive and nuanced forecasts.

However, existing ML research on market predictions has suffered several limitations. One significant drawback is that most methods require a large amount of dense historical data for training. This requirement can be resource-intensive and may not always be feasible. Additionally, these models often struggle to accurately capture market peaks and volatility, leading to less reliable forecasts during periods of rapid market changes. As such, these limitations hinder the effectiveness of stock prediction models in providing timely and precise insights, particularly under volatile market conditions.

Recently, reservoir computing (RC) has gained prominence as a streamlined yet effective neural network architecture. It has been demonstrated to accurately predict time-series data with a limited number of training samples \cite{bu2022, kumar2021efficient, rafayelyan2020large}. 
%The RC architecture is derived from various neural network architectures, including recurrent neural networks, echo state networks, and liquid state machines. 
This architecture holds significant potential as an alternative for traditional, complex neural network structures in processing financial datasets, particularly for real-time data processing when storage capacity for large datasets is limited and/or time-to-prediction is sensitive. An RC system comprises a reservoir that maps inputs into a high-dimensional space and a readout that analyzes patterns from these high-dimensional states \cite{schrauwen2007overview}. In this system, the reservoir architecture remains fixed, and only the readout is trained using simple methods such as linear regression and classification \cite{tanaka2019recent}. Therefore, reservoir computing offers fast learning and low training costs compared with other RNNs. Additionally, the static nature of the reservoir facilitates hardware implementation using a variety of optical systems, substrates, and devices. As such, optical reservoir computing has garnered increasing attention across diverse research fields\cite{schrauwen2007overview, duport2012all, van2017advances, dong2019optical, rafayelyan2020large}. These RC systems can be categorized into three main types: free-space optical RC, fiber-based temporal RC, and integrated photonic RC. In comparison, the free-space optical RC offers significant advantages in large parallelism and system optimization flexibility. 

\begin{figure*}
	\centering 	\includegraphics[width=1\textwidth]{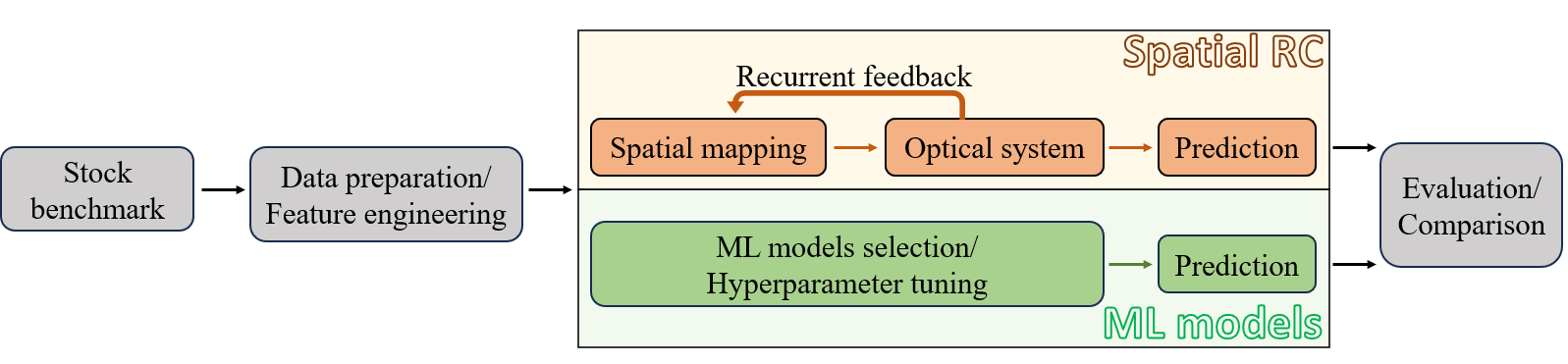}	
	\caption{Experiment diagram for stock benchmark prediction to compare spatial reservoir computing and competitive machine learning models.} 
	\label{fig:diagram}%
\end{figure*}

In this paper, we introduce and demonstrate a free-space optical RC system for predicting daily stock index prices. Utilizing a spatial light modulator (SLM), the temporal data from benchmarks is transformed into high-dimensional optical signals in both spatial and temporal domains for high-quality benchmark predictions. Owing to an innovative design of the mapping method and the large number of pixels on the modulator, the system is flexible to implement the training samples with different feature sizes simultaneously. Therefore, the information can be processed in parallel in the optical domain without compromising the time for signals' interconnection or nonlinear effect as in the temporal RC \cite{kumar2021efficient}. Furthermore, this large-scale parallelism allows multiple-step prediction, by translating temporal correlation in the time-series data to spatial coherence in the free-space optics, and conveniently encoding and reading out in the spatial domain. 

We select seven popular stock indices and gather 500 historical data points for each. Additionally, we incorporate seven macroeconomic and technical analysis features as input variables for the model training. We further optimize the optical RC system by incorporating feature correlation as weights. Our comparative analyses with popular ML pricing models indicate that our system outperforms most ML models for both one-step and multi-step predictions. Both ML and LSTM models serve as significant reference points in experimental financial research and act as essential benchmarks when compared with our optical RC experiment. Unlike previous research requiring extensive historical data, our system efficiently achieves parallel real-time predictions using no more than 500 data points with multi-dimensional features. Notably, the inclusion of feature correlation weights significantly enhances prediction performance by capturing signals and peaks instantly. The feature-weighted optimized RC demonstrates substantial potential for confidential, low-latency edge processing and practical pricing strategies.

\section{Model}

\subsection{Experiment diagram}
Figure \ref{fig:diagram} illustrates the experimental procedures described in this paper. Initially, we download historical trading datasets from stock benchmarks and perform basic data and feature engineering. Subsequently, we apply optical RC and various popular ML techniques to predict future stock index prices, respectively. We compare different machine learning models, utilizing cross-validation and hyperparameter tuning to select the ML model with the lowest error for each benchmark. Finally, we evaluate our models and compare the predictions from optical RC with the best-performing ML model for each benchmark.

\section{Datasets preparation}
This study examines seven stock indices frequently analyzed in investment strategies and portfolio management, as referenced in \cite{kumbure2022machine}. The primary datasets consist of weekday historical data from May 2023 to September 2023, sourced from Yahoo Finance. These datasets include crucial pricing details, specifically the daily closing prices. Comprehensive ticker symbols and index information are provided in Table I. The closing prices of all the stock indices are inherently noisy, non-linear, and chaotic.

%%%%%%%%%%%%%%%%%%%%%%%%%%%%%%%%%%%%%%%%%%%%%%%%%%%%%%%  indexes table
%%%%%%%%%%%%%%%%%%%%%%%%%%%%%%%%%%%%%%%%%%%%%%%%%%%%%%%
\begin{table}[h]%% placement specifier
%% Use tabular environment to tag the tabular data.
%% https://en.wikibooks.org/wiki/LaTeX/Tables#The_tabular_environment
\centering%% For centre alignment of tabular.

\scalebox{0.8}{
\begin{tabular}{|l |l |l|} 
 \hline
 \textbf{Ticker} & \textbf{Index} & \textbf{Region} \\
  \hline
% REIT & ALPS Active REIT ETF & US\\ 
VTHR & Vanguard Russell 3000 ETF & US \\
\string^N225 & Nikkei 225 & Japan\\ 
UKX.L & The Financial Times Stock Exchange 100 Index & UK \\
% \string^GSPC & S\&P 500 & US \\
\string^NYA & The NYSE Composite & US \\
\string^IXIC & NASDAQ Composite  & US \\
\string^HSI & HANG SENG INDEX & Chine \\
% \string^DJI & The Dow Jones Industrial Average & US \\
% DJCB & ETRACS Bloomberg Commodity Index Total Return ETN Series B & US \\ 
% \string^GDAXI & DAX PERFORMANCE-INDEX & German \\
MME=F & MSCI Emerging Markets Index Fut & \\

\hline
\end{tabular}
}
\caption{Global market indices used in this paper.}
\label{Table1}
\end{table}

%%%%%%%%%%%%%%%%%%%%%%%%%%%%%%%%%%%%%%%%%%%%%%%%%%%%%%%  indexes table
%%%%%%%%%%%%%%%%%%%%%%%%%%%%%%%%%%%%%%%%%%%%%%%%%%%%%%%

\normalsize

Due to the complexity of the benchmarks, a prediction model based solely on the closing price cannot gather all the necessary information for accurate predictions. Consequently, feature selection is a crucial part of the prediction enhancement. However, it is challenging to extract features from financial data. On the one hand, limiting the features can hinder the predictive model's performance. On the other hand, including all available features from the financial market can lead to a dimensionally complex and difficult-to-interpret model. Additionally, collinearity among multiple variables can negatively impact the model's performance. Some studies select the features rely solely on technical indicators \cite{park2007we, edwards2018technical, lo2000foundations, murphy1999technical}, while others incorporate historical data and macroeconomic indicators \cite{nti2020systematic, dow1997stock, chen1986economic, morck1990stock, engle2013stock}.

\begin{table}[h]%% placement specifier
%% Use tabular environment to tag the tabular data.
%% https://en.wikibooks.org/wiki/LaTeX/Tables#The_tabular_environment
\centering%% For centre alignment of tabular.

\begin{footnotesize}
\begin{tabular}{|l l c|} 

\hline
\textbf{Fundamental data} & \textbf{Source} & \textbf{Abbreviation} \\
closing price & Yahoo Finance & Close\\ 
closing price(t-1) & Yahoo Finance & Close(t-1)\\

\hline
\textbf{Macroeconomic data} & \textbf{Source} & \textbf{Abbreviation}\\

Cboe volatility index & Yahoo Finance & VIX \\
Effective Federal Funds Rate & FRED & EFFR \\
% US unemployment rate & FRED & UNRATE \\
Consumer sentiment index & FRED & UMSCENT \\
US dollar index & Yahoo Finance & DXYNYB \\
\hline

\textbf{Technical analysis data} & \textbf{Source} & \textbf{Abbreviation}\\
Moving average convergence divergence & Calculated & MACD \\
Average true range & Calculated & ATR \\
Relative strength index & Calculated & RSI \\

\hline
\end{tabular}
    
\end{footnotesize}

\caption{Features used for stock prediction in this paper.}
\label{Table2}
\end{table}

\begin{figure}[h]
    \centering
    \includegraphics[width=1\textwidth]{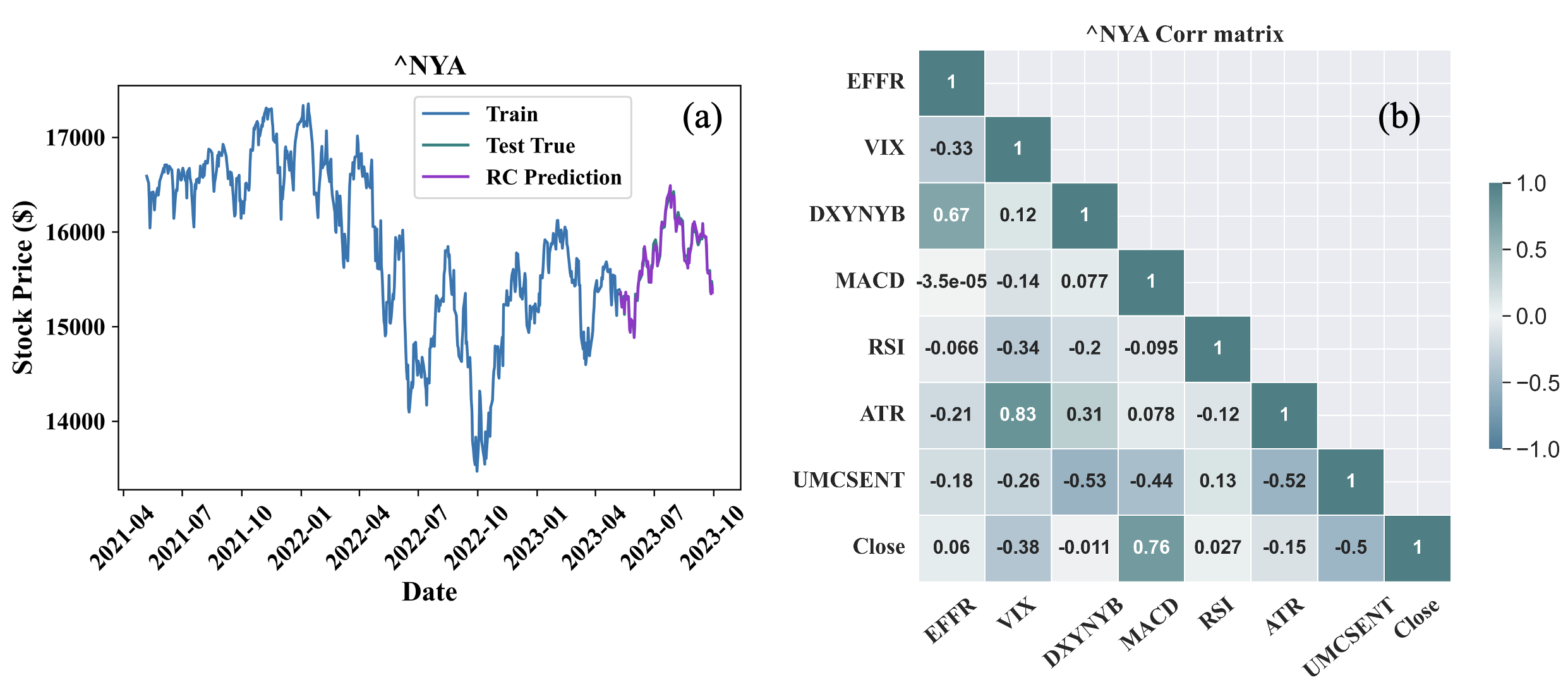}
    \caption{(a) An overview of one stock index sequence $\string^$NYA with training and testing parts. (b) A feature correlation matrix heatmap for $\string^$NYA }
    \label{fig:features}
\end{figure}

In the above considerations, we select seven main features either from macroeconomic indicators or technical indicators as the prediction inputs, as listed in Table \ref{Table2}. These features are chosen based on their high correlation with the closing price \cite{bhandari2022predicting}. As macroeconomic data can significantly impact the stock market, we consider the Cboe Volatility Index (VIX), the Interest Rate (EFFR), the Consumer Sentiment Index (UMCSENT), and the US Dollar Index (DXYNYB). The technical indicators include Moving Average Convergence Divergence (MACD), Average True Range (ATR), and Relative Strength Index (RSI). The details of the features are listed in Table \ref{Table2}. The correlation matrix among features is illustrated in Figure \ref{fig:features}(b). These values are depicted based on the intensity of the color, indicating the strength of the relationship between the given variables. In the following RC prediction, we add the bottom row of the correlation weight (CW) matrix as extra input weights \(W_{cor}\). For each index, we used 500 days as the training data and 100 days as the testing data. Each day, the training input is the previous day's closing price and the seven selected features. 

\subsection{Optical RC}
Our spatial optical RC follows the basic working principle of conventional RC, which can be described by 
\begin{equation}
    x(t)=f[\alpha x(t-1)W_{res}+\beta u(t)W_{in}], 
    \label{eq:conventional_rc}
\end{equation}
where $x(t)$ represents the reservoir state at time $t$. $f$ is the nonlinear function. $\alpha$ and $\beta$ are the feedback gain and input gain, which will be optimized for the best performance. $W_{res}$ and $W_{in}$ are the reservoir nodes' interconnection and input weight, respectively. 
Due to this updating rule of reservoir states, the current state $x(t)$ inherently contains information from the preceding state $x(t-1)$. This endows reservoir computers with the crucial property of short-term memory, a feature for effective time-series data prediction.
In the training stage, the input is injected sequentially into the reservoir, and the resulting reservoir states $\boldsymbol{x}$ for all time instances within the range [1, T] are recorded. Subsequently, the system's output weights ($W_{out}$) can be determined by minimizing the error between the predicted results ($\hat{y}=W_{out}\boldsymbol{x}$) and their corresponding true time series ($y$). The prediction performance for spatial RC and ML models are evaluated using the normalized root mean square error (NRMSE) between the true time series $y$ and the prediction results $\hat{y}$ on the testing dataset
\begin{equation}
\text{NRMSE}(y, \hat{y}) = \frac{\sqrt{\sum_{i=0}^{N - 1} (y_i - \hat{y}_i)^2}}{\sqrt{\sum_{i=0}^{N - 1} (y_i - \bar{y})^2}},
\end{equation}
where $\Bar{y}$ is the mean value of the true time series.

\begin{figure}
	\centering 	\includegraphics[width=0.6\textwidth]{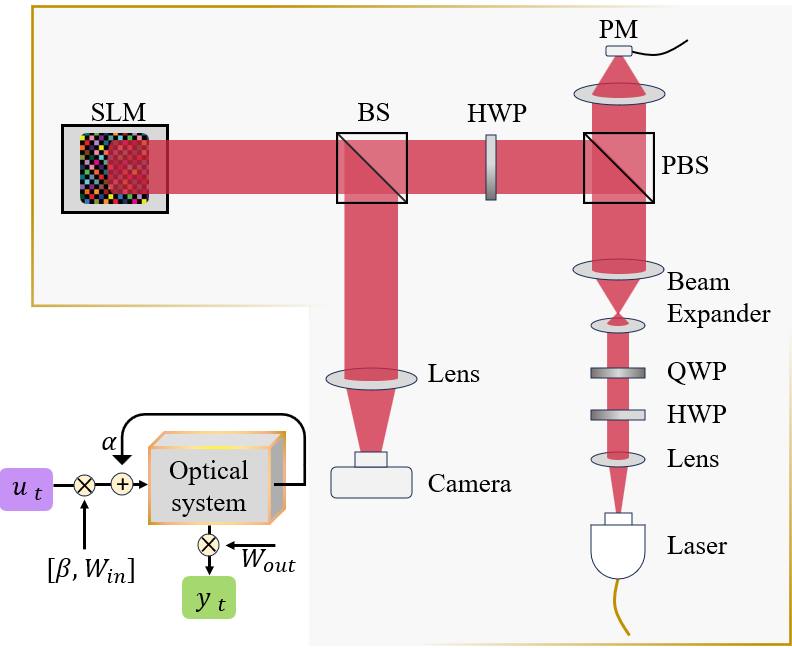}	
	\caption{The configuration of the spatial reservoir computer. SLM: spatial light modulator, BS: beam splitter, HWP: half-wave plate, QWP: quarter-wave plate, PBS: polarizing beam splitter, PM: power meter} 
	\label{fig:setup}%
\end{figure}

In our optical setup, a continuous-wave 775 nm laser diode serves as the light source, as illustrated in Fig. \ref{fig:setup}. The beam waist is adjusted to 1 mm after collimation. To match the effective area on the spatial light modulator (SLM), a 10$\times$ beam expander is used to enlarge the beam size. A power meter is implemented to capture the transmitted light passing through the polarizing beam splitter (PBS) for real-time monitoring of system stability during training and testing.
Before the beam splitter (BS), we employ a half-wave plate to rotate the light's polarization, thereby maximizing the modulation effect on the SLM. Following modulation, the light is focused, and a camera is positioned in the time-fourier domain \cite{kumar2021} to align the sensor area with the beam size for subsequent information processing. 

\begin{figure}
    \centering
    \includegraphics[width=0.8\textwidth]{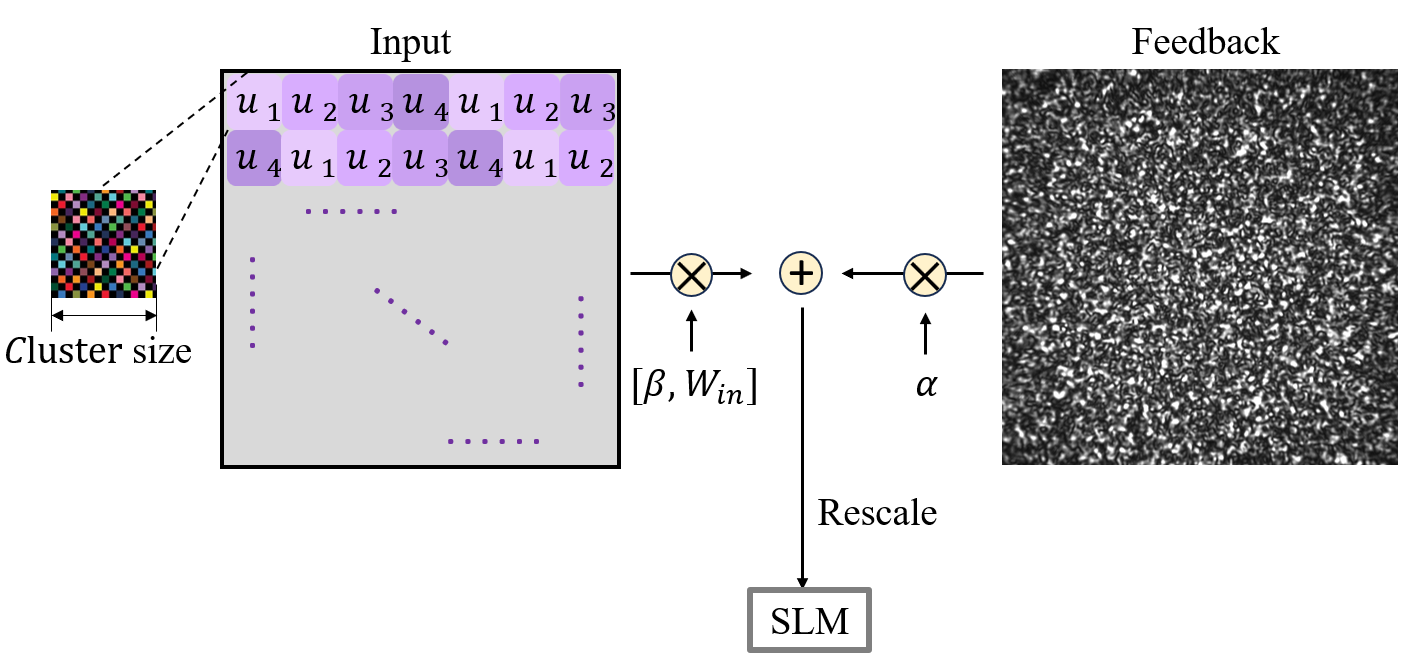}
    \caption{The mapping method to combine multi-feature inputs and the camera state for spatial light modulation.}
    \label{fig:mapping}
\end{figure}

The inset in Fig.\ref{fig:setup} illustrates the data flow diagram in our spatial RC. One key here is the combination of input values and the feedback state for spatial light modulation. Here, we use a similar mapping method as the "linear-combination mapping" in our previous paper \cite{bu2022}. The difference is with how the overall mapping area is split into several input portals. Instead of equally dividing the SLM into $n$ sections for $n$ input values, we repeat and stack the inputs until all the modulation areas on the SLM are filled. In the case where $n=4$, as shown in Fig.\ref{fig:mapping}, the 4 input values are repeated and stacked sequentially, each with a height and length of 20 SLM pixels. The mapped input is multiplied by an input scale ($\beta$) and a random input weight, while the captured camera state is rescaled with a feedback gain ($\alpha$). The combination is then linearly rescaled again to match with the $0-2\pi$ phase modulation on SLM. The nonlinearity of the system is achieved through the camera intensity reading and the intensity saturation rate. Similar to Eq.(\ref{eq:conventional_rc}), our spatial RC can be simplified as 
\begin{equation}
    x(t)=f\{[\alpha x(t-1)+\beta W_{in} \odot W_{cor} \odot u(t)]W_{res}\}.
\end{equation}
Here $W_{cor}$ is a designed weight for the features of each stock index benchmark. $f\{ \}$ and $W_{res}$ are the accumulated nonlinear effect and reservoir weight, respectively, through the whole system.

Thanks to the large number of pixels on SLM (400$\times$ 400), it can easily function as a multi-input portal, allowing a large amount of information to be encoded into the system at once. In this way, the temporal correlation among the time-series data is translated into spatial coherence of the optical field, whose dynamics implement a complex-number kernel function. Therefore, besides evaluating the 1-step prediction, where the result contains only the next step value, we also assess the system's performance for multi-step ahead prediction using more inputs. In this case, $u(t)$ is the vector containing all the feature information for different time steps. Specifically, we present examples of multi-step prediction with four-step and ten-step ahead predictions, labeled as 4-step and 10-step, respectively. Taking 4-step prediction as an example, the input vector $u(t)$ has 32 values, corresponding to one closing price and 7 features per time step. Accordingly, the prediction horizon is 4 steps containing 4 days of closing prices beyond the most recent input. This encoding method exploits the spatial correlation of the free space optics that occurs naturally in our setup.

\subsection{Multiple ML and LSTM models}
This paper uses LSTM and various regression methods to predict each benchmark \cite{bhandari2022predicting, sangiorgio2020robustness}. The LSTM architecture includes an internal memory that acts like a local variable, storing information as it processes a sequence of inputs. It uniquely handles outputs by recycling the production from one time-step as the input for the subsequent time-step\cite{hochreiter1997long}. This recycling process enables the LSTM to make informed predictions by considering both the current input and the output from the immediate preceding timestep. Besides LSTM, we also compared multiple regression models for stock index prediction. The regression methods used contain Lasso, ElasticNet, RandomForestRegressor, LinearRegression, Ridge, SGDRegressor, KNeighborsRegressor, DecisionTreeRegressor, and BayesianRidge\cite{li2014forecasting, basak2019predicting, qian2007stock}. 

For multi-step prediction, the data preparation for the time-series sequence with features follows the same approach as in optical RC in the previous subsection. The regression methods remain consistent with the one-step prediction, while the model dimensions in the LSTM are adjusted to align with the prediction horizon.

\section{Results and analysis}

\begin{figure}
    \centering
    \includegraphics[width=1.0\textwidth]{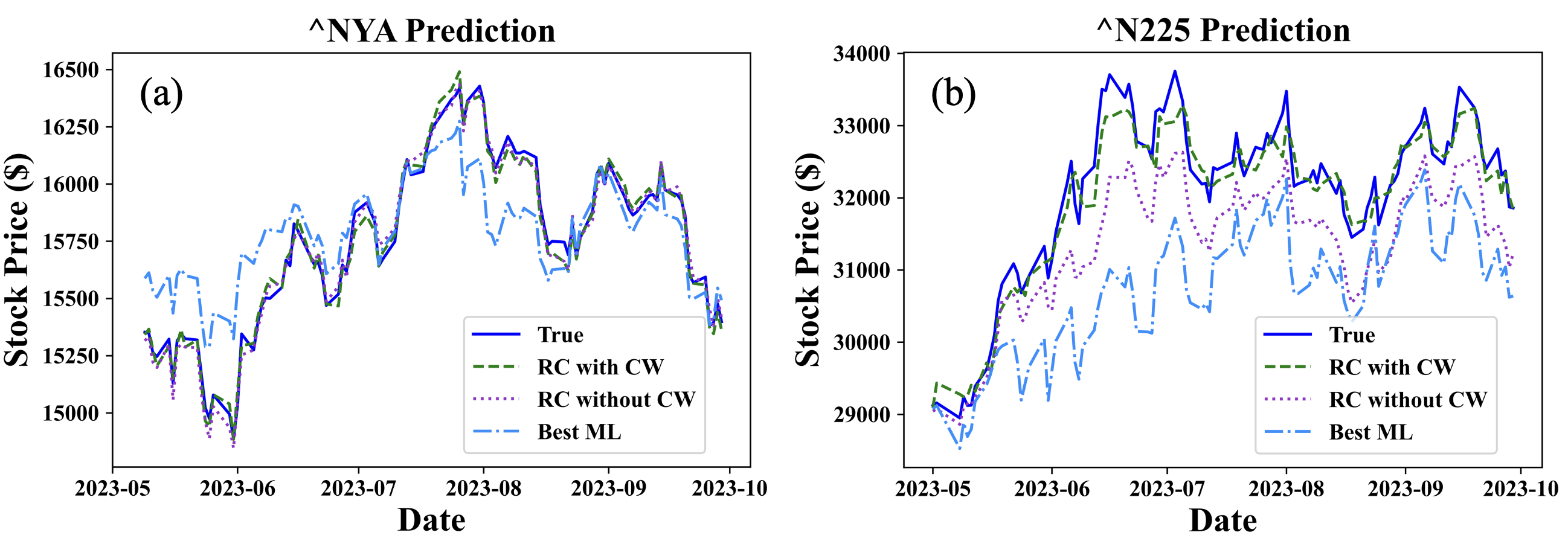}
    \caption{True values versus 1-step prediction results from optical RC with and without feature correlation weights, contrast to the best-performing ML results, shown for $\string^$NYA (a) and $\string^$N225 (b).}
    \label{fig:pred1}
\end{figure}

Figure \ref{fig:pred1} (a) and (b) are two examples of 1-step prediction results for stock indices $\string^$NYA and $\string^$N225, respectively, showing the ground truth, the prediction results from RC with and without CW. Also shown are the prediction results of the best ML model, which is the Lasso regression model for both $\string^$NYA and $\string^$N225. 
The NRMSEs of RC prediction for $\string^$NYA with and without CW are 0.104 and 0.108, respectively, whereas the best ML regression result is 0.558, which is over five times worse than the RC prediction. To compare the overall prediction performance with the results for $\string^$N225 in Fig.~\ref{fig:pred1}(b), the prediction results for $\string^$NYA show a small difference between predictions with and without CW.
The difference for $\string^$N225 between the two is much larger, as illustrated in Fig.\ref{fig:pred1}(b), where the NRMSE is 0.688 without CW, and only 0.236 with CW. This indicates that the added feature correlation weights can significantly improve the prediction performance for this benchmark.

In contrast, the predictive performance of ML models is considerably inferior to that of the RC, with the most accurate ML model achieving an NRMSE of 1.309 for $\string^$N225. This is clear in Fig. \ref{fig:pred1}, where both the best ML curves for $\string^$NYA and $\string^$N225 in Fig. \ref{fig:pred1} (a) and (b) lag behind the price movement trend and deviate significantly from the true values. Conversely, our optical RC predictions capture the stock volatility and nonlinear trends more accurately. These predictions align better with the real-time series stock prices than the best ML regression predictions.

\begin{figure}
    \centering
    \includegraphics[width=0.7\textwidth]{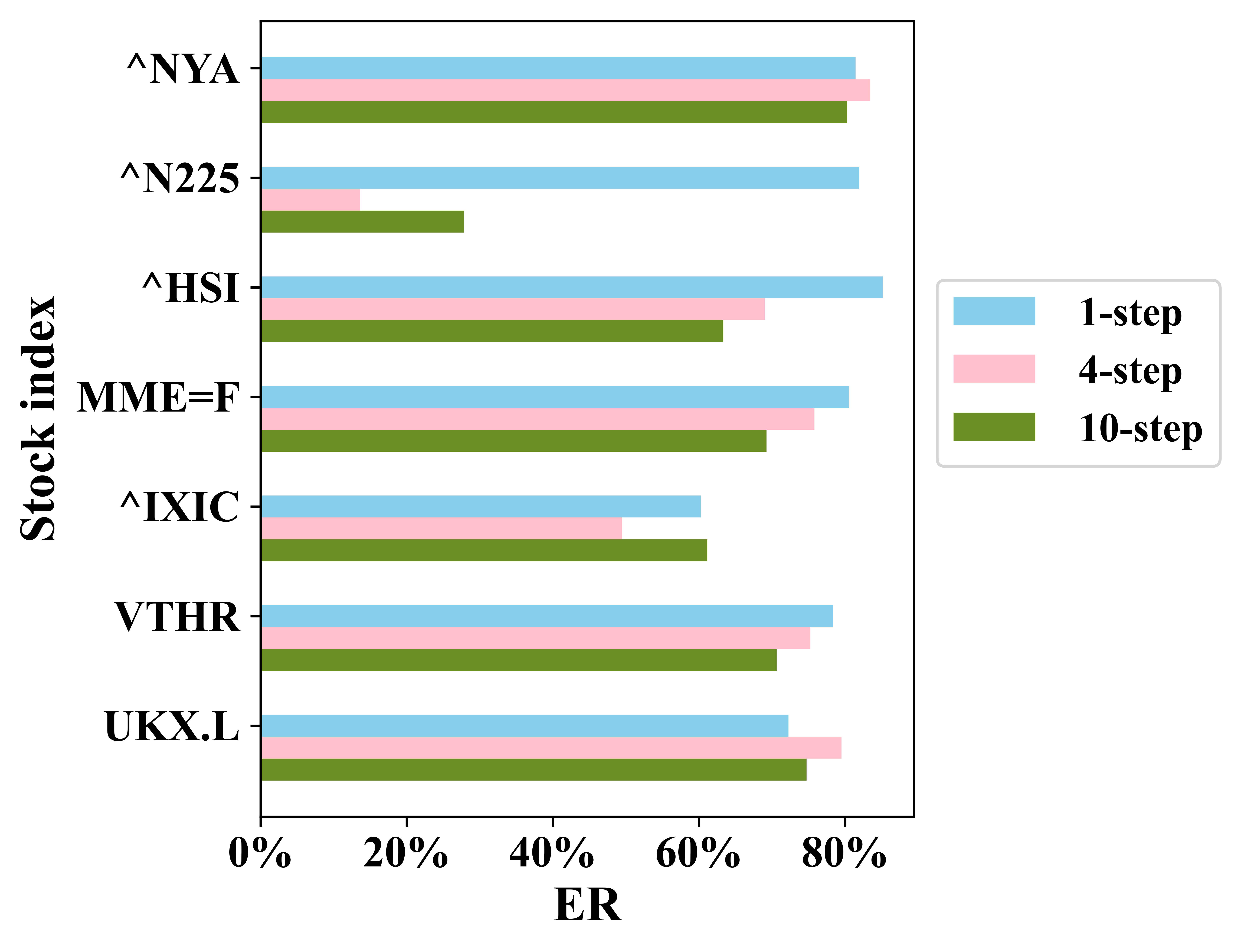}
    \caption{The error reduction between results from the best ML and RC models for different stock indices.}
    \label{fig:error}
\end{figure}

To quantify their performance difference, Figure \ref{fig:error} lists the error reduction (ER) between the best ML model and optical RC predictions for seven stock indices. The ER is calculated using the formula 
\begin{equation}
\textrm{ER}=1- \frac{\textrm{NRMSE}_{RC}}{\textrm{NRMSE}_{ML}},
\end{equation}
where $\textrm{NRMSE}_{RC}$ and $\textrm{NRMSE}_{ML}$ are the NRMSE of the RC and the best-performing ML methods, respectively. 
The above ER measures the performance improvement by RC compared to the best ML model. A near 100\% ER indicates negligible relative error achieved by RC.  
As shown, the RC model significantly outperforms the ML model, with all ER values being positive and more than half of the results exceeding 70\% for the different stock indices. Interestingly, for the stock indices $\string^$N225, $\string^$HSI, MME$=$F, and VTHR, the RC model shows superior performance in 1-step predictions compared to multi-step predictions. For other indices, the RC model achieves comparable or higher ER values in multi-step predictions than in 1-step predictions.

\begin{figure} 
    \centering
    \includegraphics[width=1.0\textwidth]{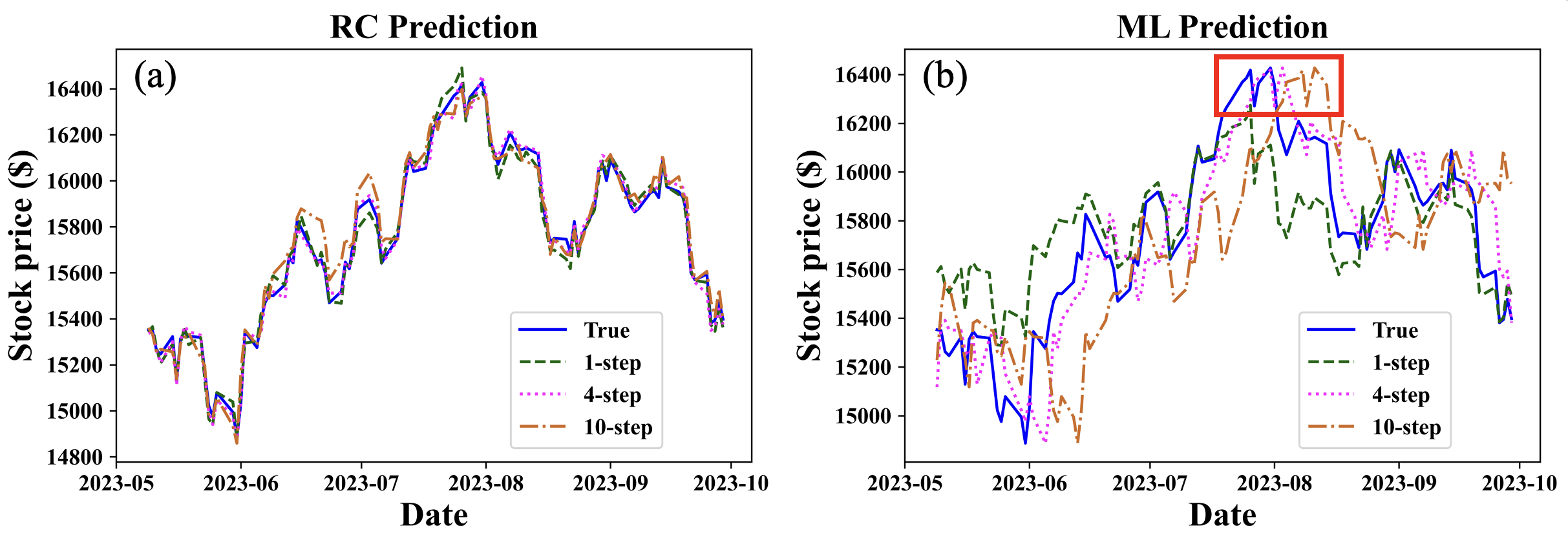}
    \caption{1-step and multi-step predictions for stock index $\string^$NYA by (a) RC and (b) ML models  }
    \label{fig:rc_vs_reg}
\end{figure}

To gain insights into prediction performances between RC and ML models, we plot the prediction results of the two models for $\string^$NYA in Fig. \ref{fig:rc_vs_reg}(a) and (b), respectively. 
Fig.7(a) shows the prediction outcomes from our RC model. The different-step prediction lines fit closely to the ground truth, with only a slight offset around a few peaks for 10-step prediction. 
However, the results of the ML model results in Figure \ref{fig:rc_vs_reg}(b) are more dispersed to compare the results for RC prediction. Additionally, the red rectangular area in Figure \ref{fig:rc_vs_reg}(b) highlights a 'delayed' effect in the multi-step prediction. Larger prediction dimensions result in greater drift from the ground truth. Consequently, as the forecasting horizon extends, prediction accuracy diminishes, leading to greater deviations from actual outcomes.

%%%%%%%%%%%%%%%%%%%%%%%%%%%%%%%%%%%%%%%%%%%%%%%%%%%%%%%  data source table
%%%%%%%%%%%%%%%%%%%%%%%%%%%%%%%%%%%%%%%%%%%%%%%%%%%%%%%

\section{Conclusion}

In conclusion, we have established an optical reservoir computing system in free space that effectively captures broader market behavior and demonstrates superior performance in stock index prediction. Compared to the existing machine learning methods and long short-term memory neural networks, our system offloads intense computation to an optical system with low energy consumption, showing significant advantages in handling high volatility and nonlinear market behaviors using limited data. This method promises a cost-effective approach to real-time, parallel, multi-dimensional data processing and predictions, with broad applications in financial forecasting. The next steps include moving more data processing into the optical domain, for high-speed, high-quality, and versatile applications in financial analysis.

% \appendix
% \section{Example Appendix Section}


% %% For citations use: 
% %%       \cite{<label>} ==> [1]

% %%
% Example citation, See \cite{lamport94}.

\bibliographystyle{elsarticle-num} 
\bibliography{main.bib}




% %% For numbered reference style

%   \textit{\LaTeX: a document preparation system},
%   Addison Wesley, Massachusetts,

% \end{thebibliography}
\end{document}